\documentclass[sigconf,screen,nonacm]{acmart}
\AtBeginDocument{%
  \providecommand\BibTeX{{%
    \normalfont B\kern-0.5em{\scshape i\kern-0.25em b}\kern-0.8em\TeX}}}

\usepackage[ruled,linesnumbered]{algorithm2e}

\setcopyright{acmlicensed}
\copyrightyear{2018}
\acmYear{2018}
\acmDOI{XXXXXXX.XXXXXXX}

\acmConference[Conference acronym 'XX]{Make sure to enter the correct
  conference title from your rights confirmation emai}{June 03--05,
  2018}{Woodstock, NY}
\acmISBN{978-1-4503-XXXX-X/18/06}

\begin{document}

\title{Improving Interpretable Embeddings for Ad-hoc Video Search with Generative Captions and Multi-word Concept Bank}

\author{Jiaxin Wu}
\email{jiaxin.wu@my.cityu.edu.hk}
\affiliation{%
  \institution{City University of Hong Kong}
  \country{Hong Kong}
}

\author{Chong-Wah Ngo}
\email{cwngo@smu.edu.sg}
\affiliation{%
  \institution{Singapore Management University}
  \country{Singapore}
}

\author{Wing-Kwong Chan}
\email{wkchan@cityu.edu.hk}
\affiliation{%
  \institution{City University of Hong Kong}
  \country{Hong Kong}
}

\renewcommand{\shortauthors}{Trovato and Tobin, et al.}

\begin{abstract}
Aligning a user query and video clips in cross-modal latent space and that with semantic concepts are two mainstream approaches for ad-hoc video search (AVS). However, the effectiveness of existing approaches is bottlenecked by the small sizes of available video-text datasets and the low quality of concept banks, which results in the failures of unseen queries and the out-of-vocabulary problem. This paper addresses these two problems by constructing a new dataset and developing a multi-word concept bank. Specifically, capitalizing on a generative model, we construct a new dataset consisting of 7 million generated text and video pairs for pre-training. To tackle the out-of-vocabulary problem, we develop a multi-word concept bank based on syntax analysis to enhance the capability of a state-of-the-art interpretable AVS method in modeling relationships between query words. We also study the impact of current advanced features on the method. Experimental results show that the integration of the above-proposed elements doubles the R@1 performance of the AVS method on the MSRVTT dataset and improves the xinfAP on the TRECVid AVS query sets for 2016--2023 (eight years) by a margin from 2\% to 77\%, with an average about 20\%. 


\end{abstract}



\keywords{Ad-hoc video search, Interpretable embedding, Large-scale video-text dataset, Concept bank construction, Out of vocabulary} 



\maketitle

\section{Introduction}
With the ever-growth of video data, e.g., videos sharing on YouTube or TikTok, text-to-video search is an essential tool for users to find videos of their interests. Especially, ad-hoc video search (AVS), which allows users to retrieve videos through an open-vocabulary textual query, has attracted lots of attention in many years \cite{trecvid2006,Informedia2018,Waseda2018, w2vvpp,wu2023ITV}. AVS is challenging as it does not provide any annotated data for training, and the test collection is usually huge (e.g., V3C1 dataset \cite{V3C1}, which has over 1 million video clips).

Concept-based search \cite{Waseda2018,Boer:Semantic,QKR:AVS} and embedding-based search \cite{RUCMM_2018,w2vvpp, hgr, dualconding} are two mainstream approaches for ad-hoc video search. Concept-based search indexes a user query and video clips by semantic concepts. Specifically, the concepts of the video clips are extracted by classifiers trained on image/video classification datasets. The query text is mapped to concept tokens of a concept bank through a concept selection. The effectiveness of a concept-based search relies on the qualities of the concept selection and the concept bank, as well as the accuracies of the video concept extractors. In contrast, by getting rid of the tedious concept extraction and selection, embedding-based search (aka concept-free search) aligns a user query with video clips in a joint latent space. With more video caption datasets available, embedding-based search has outperformed concept-based search and has become a mainstream approach for AVS since 2018. However, since the alignment is conducted on the high-dimensional latent space, the result of the embedding-based search is not explainable and predictable. To this end, the dual-task model \cite{dual_task} proposes to interpret the joint latent space with semantic concepts and align a user query with video clips in an interpretable embedding space. Some recent studies \cite{RIVRL,wu2023ITV} follow the effort of the dual-task model to build interpretable approaches. For example, the ITV model \cite{wu2023ITV} proposes more consistent interpretations for query-video embeddings with unlikelihood training, and the RIVRL model \cite{RIVRL} aligns a user query and videos in both a latent space and a concept space. As concept-based search and embedding-based search are complementary, the interpretable approaches are promoted as a new state-of-the-art for the AVS task. 

Nevertheless, the existing AVS approaches are bottlenecked by small video-text datasets for training and usually fail on unseen queries. On the one hand, some approaches try to take advantage of large-scale image-text datasets and utilize text-to-image retrieval models for text-to-video retrieval \cite{Waseda2022,Waseda2023}. To some extent, the text-to-image retrieval models work for small-size datasets, e.g., MSRVTT dataset \cite{msr-vtt} with 10k videos. However, when the dataset size becomes huge, such as over 1 million video clips, either text-to-image retrieve models or their extension to video level, e.g., CLIP \cite{CLIP} or CLIP4CLIP \cite{Luo2021CLIP4Clip}, is not competitive to a state-of-the-art AVS model trained on small-size video caption datasets as evidenced in the TRECVid AVS evaluations \cite{trecvid2020,Trecvid2021,trecvid2023}. On the other hand, some huge text-video datasets (e.g., HowTo100M \cite{miech19howto100m}, and WebVid \cite{Bain21_frozenInTime}) are released to facilitate the learning of text and video alignments. However, existing large-scale datasets only have one caption per video, which overlooks rich semantics in videos. As a query text is usually ad-hoc, having multiple versions of video captions is essential for learning robust query-video alignments. In addition to the small-size dataset problem, the existing approaches only leverage the word-only concept banks for interpretation \cite{wu2023ITV,RIVRL}. The word-only concept bank is inherently limited in modeling the word-to-word relationship, resulting in out-of-vocabulary in modeling word composition, such as those with prepositional words (e.g., "in front of").

In this paper, we follow the effort of the interpretable approaches and propose three general and feasible components to address the dataset size and out-of-vocabulary problems: (1) Capitalizing on a generative model, we construct a new large-scale text-video dataset automatically for video-text retrieval models without human effort. The dataset (named WebVid-genCap7M) has 7 million generated text-video pairs, which is available online\footnote{\href{https://drive.google.com/file/d/18Dh20_ZlSGJ_XAFM2P5dpd3qSIR-vSBJ/view}{WebVid-genCap7M dataset}}, and it is used for the retrieval model pre-training. (2) We develop a multi-word concept bank to address the out-of-vocabulary problem in concept-based search by incorporating various phrases based on syntactic analysis of sentences. The evaluation shows that the new concept bank is able to bring an average of about 60\% gain of xinfAP to a state-of-the-art concept-based search on TRECVid AVS query sets, especially on those out-of-vocabulary queries. (3) We also investigate the impact of recent advanced text/video features on a state-of-the-art AVS model. The experimental results demonstrate that integrating the three newly introduced components into the AVS model outperforms most of the top-1 results reported on the TRECVid AVS benchmarks over the past eight years, contributing to a new state-of-the-art performance in the AVS task.

\section{Related work}
\subsection{Ad-hoc Video Search}
Ad-hoc video search, a task that has been consistently evaluated yearly in TRECVid, has its origins dating back to 2003 \cite{trecvid2006}. Given a query (i.e., a textual description of the desired search content), the search engine needs to process the query and return a ranked list of video clips \cite{Trecvid2016}. 

From its very beginning, the mainstream approaches focus on using semantic concepts to tackle the task \cite{Snoek:concept_video_retrieval}, i.e., concept-based search. Early efforts are devoted to concept bank development and ontology reasoning \cite{Naphade:LSCOM,Jiang:semantic_concept,Snoek:challenge}, and concept screening, representation and combination \cite{ITI-CERTH2018,NII2016,Lu:ICMR,wasedaAVS2020,WasedaMeiseiSoftbank2019,Informedia2018}. Out-of-vocabulary (OOV) is one of the challenges in concept-based search. Existing approaches mainly address the challenge in two ways: by building a large concept bank or by having a good strategy for concept selection. The most recent methods concentrate on building a large concept bank, e.g., over 47,000 concepts  \cite{wasedaAVS2020,WasedaMeiseiSoftbank2019}. Besides, Nakagome et al. \cite{Waseda2018} select concepts for a query by applying ontology analysis on WordNet \cite{wordnet} to find hypernyms for query tokens and measure semantic similarity between concepts and the hypernyms. Huang et al. \cite{Informedia2018} also perform a series of measurements between query terms and concepts to tackle the OOV problem, such as synset similarity based on WordNet taxonomy and explicit semantic analysis based on Wikipedia. Although numerous efforts have been made \cite{Waseda2018,Informedia2018,Boer:Semantic,Jiang:semantic_gap,vireoAVS2019}, it still has difficulty in automatically selecting concepts for queries. Human intervention is always needed in picking concepts to boost AVS performances \cite{WasedaMeiseiSoftbank2019,wasedaAVS2020}. Concept-based search performs well on those queries when their information need could be precisely identified by a list of concepts. However, the inherent problem of existing approaches is expression ambiguity. For example, it is hard to precisely convey the search ambition by using a word-only concept bank with the concepts \lq\lq hold\rq\rq, \lq\lq hand\rq\rq, and \lq\lq face\rq\rq\ for the query \textsl{Find shots of a person holding his hand to his face}, especially if a query involves prepositional words. To this end, we propose a multi-word concept bank in this paper to enhance the concept-based approach in modeling relationships between query words. Different from the existing approaches, which accumulate concepts from existing object, action, and scene classification datasets to build a concept bank, our multi-word concept bank is built based on syntax analysis of video captions. In addition to nouns and verbs, we also extract five common phrases in English (i.e., noun phrases, verb phrases, adjective phrases, prepositional phrases, and quantifier phrases) as concepts. As a result, the previous query can be identified by a verb phrase with a prepositional phrase, i.e., "hold hand" and "to face".

Embedding-based search, which encodes a user query and video clips in a high-dimensional latent space, is another mainstream approach in AVS. With more video captioning datasets (e.g., MSRVTT \cite{msr-vtt}), TGIF \cite{tgif}, VATEX \cite{VATEX}) available,  embedding-based search has significantly outperformed concept-based search since 2018. Many models have been applied for AVS, including VideoStory \cite{videostory}, visual semantic embedding (VSE++) \cite{vse}, intra-modal and inter-modal attention networks  (IAN) \cite{Informedia2018}, Word2VisualVec (W2VV) \cite{w2vv}, dual encoding \cite{dualconding}, HGR \cite{hgr} and SEA \cite{tmm2021-sea}. The differences between these models are how they encode and represent a query text. For example, VSE++ embeds the query by a recurrent network \cite{vse}, while IAN assigns different attention to the query \cite{Informedia2018}. A more complex text encoder is used in W2VV \cite{w2vv}, which puts bag-of-words (BoW), word2vec (W2V), and word sequence altogether. The extension of W2VV (W2VV++) is the first model that significantly outperforms concept-based search on AVS \cite{Trecvid2018}. \cite{W2VVpp_Case_Study} studies some variants of W2VV++ by reducing or replacing networks of the text encoder. Building on top of W2VV++, a recent dual encoding network \cite{dualconding} reports better performances by a multi-level text assembler. Three different encoders are applied, including a short-term local encoder, a word pooling encoder, and a word sequence encoder. Dual coding has some variants. For example, Damianos et al. extend it by adding an attention mechanism \cite{dual_coding_improved}. Besides, inspired by the recent progress of graph convolutional networks, HGR \cite{hgr} encodes a query by various graphs. The most recent works, SEA \cite{tmm2021-sea} and LAFF \cite{LAFF}, design query assemblers of several encoders and train the encoders in multiple spaces with multiple losses. They report state-of-the-art performances on TRECVid query sets. Nevertheless, the result of the embedding-based search is not interpretable and predictable. For instance, two queries \textit{Find shots of a bald man} and \textit{Find shots of a hairless man}, although they have similar information needs, the results are dramatically different. 

A hybrid of concept-based search and embedding-based search has also been explored  \cite{mediamill2017,Waseda2018,Informedia2018,WasedaMeiseiSoftbank2019, vireo2017,EURECOM2019,Kindai_kobe2019, wu2023ITV}. The early studies include the fusion of VideoStory embedding and concept features \cite{mediamill2017}, leading to a boost over individual models. In the recent TRECVid evaluation, lately fusing concept-based search and embedding-based search has become a norm \cite{Waseda2018,Informedia2018,WasedaMeiseiSoftbank2019,vireo2017,EURECOM2019,Kindai_kobe2019,wasedaAVS2020}. Although they are shown to be complementary, the fact that both searches are produced by two individual models trained with different forms of data has tremendously increased the system complexity. To this end, the dual-task model \cite{dual_task} proposes to train them with the same training set of data in an end-to-end manner via dual-task learning. Specifically, it learns video-query features in a joint space and interprets the space with semantic concepts to solve the interpretation problem of the embedding-based search. Following the effort of the dual-task model, RIVRL \cite{RIVRL} aligns a query and video clips in a feature space and a concept space, and ITV \cite{wu2023ITV} proposes unlikelihood training to complement the likelihood training in \cite{dual_task} to have more consistent query-video embedding interpretations. As the concept and embedding-based searches are complementary, this kind of interpretable approach is a new state-of-the-art for the AVS task. In this paper, we follow the effort and propose three components to enhance an interpretable approach.

\subsection{Large-scale Video-Text Datasets}
In recent years, we have witnessed large-scale text-image datasets in the open domain, such as WebImageText \cite{CLIP} with 400M text-image pairs and LAION-5B \cite{schuhmann2022laion5b} with 5B text-image pairs, have substantially improved the performances of multiple text-image tasks, e.g., text-image retrieval and image captioning. Similarly, large-scale video-text datasets are proposed to facilitate the progress of video-text tasks. For example, the HowTo100M dataset \cite{miech19howto100m}, which has 136M video clips covering 23k human activities and each video clip has a caption obtained from a narration, has promoted the progress of the text-based action location task and text-to-video retrieval in instruction domains such as cooking. However, as the caption is a narrative (a subtitle), it cannot be guaranteed that it is aligned with the visual content. To this end, another large-scale text-video dataset (i.e., WebVid2M \cite{Bain21_frozenInTime}) is recently proposed for the open domain. It has 2.5M video clips, and each of them is associated with a manually annotated caption. However, the caption lengths and styles have big differences in this dataset. The caption lengths range from 4 to 40.  Some of the captions have well-defined sentence formats, while some have less defined formats, such as keywords and disjoint sentences. Moreover, the existing large-scale video-text datasets only have one caption per video clip, where the rich semantic information of a video is overlooked. In this paper, we follow the successful attempts at generating captions for images, such as LAION-COCO\footnote{https://laion.ai/blog/laion-coco/} and ALIP \cite{ALIP}, and construct a large-scale video-text dataset (WebVid-genCap7M), by automatically generating captions for videos. Specifically, we generate multiple captions with diverse content and different wordings for a video clip to cover the rich information in semantic and to facilitate the learning of robust alignments between a video clip and different wording sentences.

\subsection{Concept Bank Construction}
A good quality concept bank is effective in addressing the out-of-vocabulary (OOV) problem in concept-based search. The existing concept-based approaches usually construct a concept bank by composing various classes of multiple off-the-shelf classification datasets. For example,  \cite{Informedia2018} constructs a concept bank by accumulating object classes from YFCC100M \cite{YFCC100M}, action classes from UCF101 \cite{UCF101} and Kinetics \cite{kinetics600}, location classes from Place365 \cite{place365}, and a combination of person+object+action+location classes from SIN346 \cite{TRECVid2014}. However, the current way of constructing a concept bank inherently limits the ability to model a search intention with a focus on relationships and attributes. Although
\cite{WasedaMeiseiSoftbank2019, wasedaAVS2020} construct ATTRIBUTES300 and RELATIONSHIPS53 based on the Visual Genome dataset \cite{Visual_Genome} to better model the attributes and relationships between persons and objects, the number of the concept and the fact that the concept extractors are trained on images restrict their effectivenesses on ad-hoc queries and large video corpus. Recently, the interpretable approaches \cite{dual_task,wu2023ITV,tpami21-dual-encoding, RIVRL} built a concept bank by directly dividing a sentence into concept tokens. Although the concept bank has all the basic elements of a sentence (e.g., nouns, verbs, adjectives), they are word-only concepts. In this paper, we propose to build a multi-word concept bank based on syntax analysis of sentences to include both words and phrases. The experimental results show that the newly constructed concept bank, which includes the basic elements of a query sentence and the possible relationships of query words, has significantly improved the performance of a concept-based search, and the performance improvements are significant on the OOV queries. Moreover, the new concept bank, along with the new dataset and recent-advanced features, promotes the concept-based search as being competitive or even better than embedding-based search on some TRECVid AVS query sets.

\section{Improved Interpretable embeddings}

In this section, we illustrate how to plug in three proposed components to a state-of-the-art interpretable embedding model (ITV)~\cite{wu2023ITV}. Our proposed model is the ITV model with all three components (named improved ITV).

Given a video $v$ and a text query $q$, an interpretable embedding model encodes them by a visual encoder $H(x)$ and a textual encoder $F(x)$ to a joint embedding space as $H(v) \in \mathbb{R}^{d}$ and $F(q) \in \mathbb{R}^{d}$, respectively where $d$ is the dimension of the latent space. The video/text embedding is subsequently interpreted by a concept decoder $G(x)$ and outputs a probability vector over $n$ concepts, e.g., $G(H(v))$ = $[p_1,p_2,...,p_n]$ where $p_i$ indicates the probability of the concept $i$ being present in the video $v$. The interpretable embedding model is trained end-to-end via dual-task learning. On the one hand, visual and textual encoders are trained to ensure text-video pairs stay close in the joint space. On the other hand, the concept decoder is trained to decode concepts from a visual/textual embedding for describing the semantics in the text (video captions). 
In the inference stage, concept-based search aligns a query text with videos based on the similarity of $G(F(q))$ and $G(H(v))$ while embedding-based search is based on $F(q)$ and $H(v)$. Fusion search combines concept-based search and embedding-based search by a linear function. 

\begin{table*}[]
\caption{Dataset statistics of existing video caption dataset on the open domain. We automatically create a new large-scale video-caption dataset named WebVid-genCap7M with multiple captions per video.}
\label{dataset_info}

    \centering
\begin{tabular}{lcccccc}
\toprule

dataset      &domain  & caption type    & \#videos & \#captions & \#avg token & \#cap/video \\
\midrule
MSRVTT \cite{msr-vtt}      & open               &manually annotated                     & 10K     & 100K      &  9.28               & 10                 \\
TGIF \cite{tgif}        & open               &manually annotated                     & 100K    & 128k      &  11.28                   & 1              \\
VATEX \cite{VATEX}        & open               &manually annotated                     & 34.9k        & 349k         &  15.23            & 10                  \\
HowTo100M \cite{miech19howto100m}       & instruction        &subtitles                 & 136M        &   136M         &  4.16                   &  1                  \\
WebVid2M \cite{Bain21_frozenInTime}    & open               & manually annotated                         & 2.5M    & 2.5M      & 12                    & 1     \\
WebVid-genCap7M & open               &automatically generated                 & 1.44M    & 7.1M        &   9.94                  &  5\\       \bottomrule           
\end{tabular}
\end{table*}

\begin{figure}
    \centering
    \includegraphics{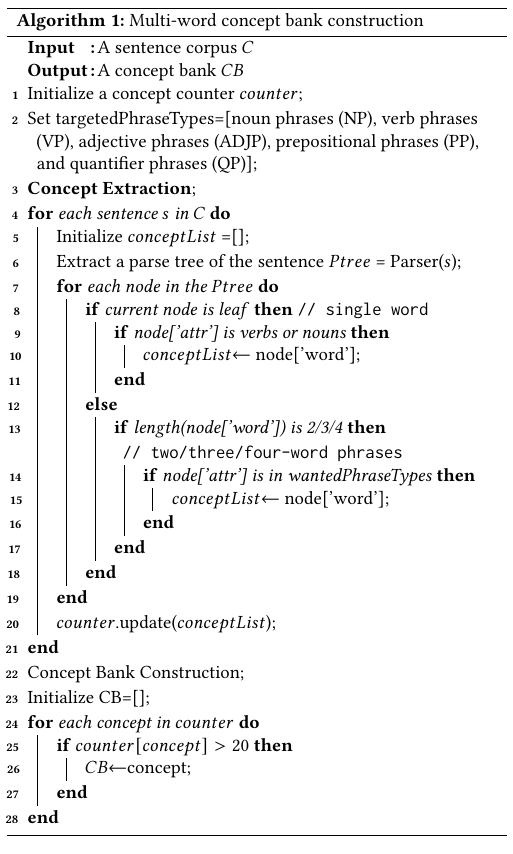}
    \label{fig:algorithm}
\end{figure}

\subsection{Multi-word Concept Bank}
Existing interpretable models \cite{dual_task,wu2023ITV,tpami21-dual-encoding} construct a concept bank by directly dividing text into individual word tokens without syntax analysis, and interpret embeddings with only words. Constructing a concept bank in such way neglects composition/relationship between words in a sentence, which leads to incomprehensible and imprecise interpretation. For example, interpreting the embedding of a compositional phrase (e.g., "sign language") with individual word concepts (e.g., "sign" and "language") could mislead the model understanding. Interpreting the embedding of an adjective phrase (e.g., "a blue shirt") with object/being and attribute words separately (e.g., "blue" and "shirt") will eventually lead to an imprecise interpretation. Especially, when there are multiple adjectives in a sentence, simple word tokenization will result in various mismatches of attributes and objects/beings.

In this paper, we propose to perform syntax analysis on text before building a concept bank and associate embeddings with both word and phrase concepts to provide a more comprehensive and precise interpretation. The process of building a multi-word concept bank is illustrated in Algorithm 1. It has two stages: concept extraction  and concept bank construction. For each sentence in the corpus, we extract word and phrase concepts. Specifically, given a sentence, we first perform syntax analysis and obtain its parse tree. Next, the algorithm goes through each node of the parse tree to extract nouns, verbs, and five main types of phrases in English: noun phrases, verb phrases, adjective phrases, prepositional phrases, and quantifier phrases. After concept extraction, we count the frequency of a concept that appears in the corpus, and those concept words or phrases that appear more than 20 times are used to build the concept bank.

\subsection{Large Video-GenText Dataset for Pre-training}

\begin{figure*}[]
    \centering
    \includegraphics[width=1\linewidth]{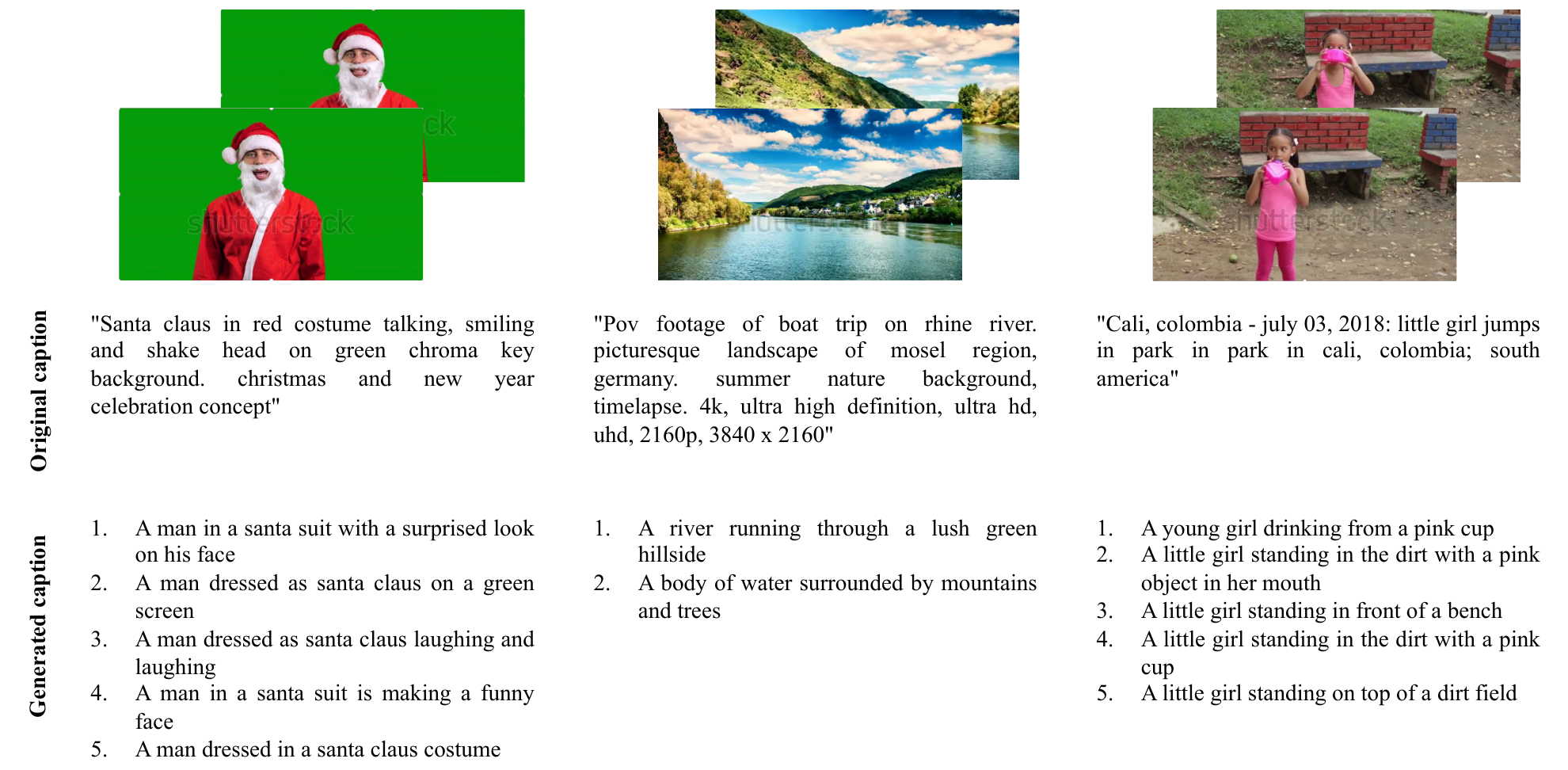}
    \caption{Example video-GenCaption pairs from the WebVid-genCap7M dataset along with the original captions in WebVid2M~\cite{Bain21_frozenInTime}. }
    \label{fig:webvid-genCap_examples}
\end{figure*}

To learn effective and representative cross-modal interpretable embeddings, having sufficient annotation data is essential. Furthermore, an image is worth a thousand words, and a video is a series of images. Having multiple captions for a video is important to learn robust alignments between a video and sentences. However, the existing video caption datasets are either small size (e.g., MSRVTT~\cite{msr-vtt}, TGIF \cite{tgif}, VATEX \cite{VATEX}) or have a small number of captions per video (e.g., HowTo100M \cite{miech19howto100m}, WebVid2M \cite{Bain21_frozenInTime}) as shown in Table \ref{dataset_info}. Generating pseudo labels for unlabeled data in supervised learning has shown effectiveness in improving the performance \cite{pseudo_labelICMLW}, and there are some successful attempts at generating synthetic captions for image dataset \cite{ALIP} to improve text-image retrieval. Following these efforts and capitalizing on the recent progress of pre-trained generative models, we explore generating captions for videos to supervise the training of cross-modal interpretable embeddings. 

In this paper, we generate synthetic captions for videos\footnote{We only manage to download 1.44M videos based on the official provided video URL.} in the WebVid2M dataset \cite{Bain21_frozenInTime} to pre-train the interpretable embedding model. Specifically, for each video, we extract around five frames, and the duration of two adjacent frames is around 3.6 seconds to ensure visual difference. For each frame, we generate a caption using an image captioning model \cite{li2022blip}. Eventually, a video is associated with multiple generated captions for cross-modal interpretable embedding learning. In total, we generate seven million captions for about 1.44 million videos. We name this dataset WebVid-genCap7M. As shown in Table \ref{dataset_info}, compared to the existing dataset, WebVid-genCap7M has a larger scale and a larger number of captions per video than other huge datasets. By having denser frame sampling or having more captions for a frame, the number of captions can become larger. Figure \ref{fig:webvid-genCap_examples} shows some example video-caption pairs in WebVid-genCap7M along with the original video captions in WebVid2M \cite{Bain21_frozenInTime} for comparisons. The captions of WebVid2M \cite{Bain21_frozenInTime} have large differences in length and format. For example, the lengths of the original caption in the second and third example in Figure \ref{fig:webvid-genCap_examples} are 29 and 17, respectively. They also have different formats. For example, the first original caption is in sentence format, while the third is in keyword format. In contrast, our generated captions are similar in length and have well-defined sentence structure. Moreover, they are diverse in wording expressions and mention extra details that are correct for the video but not mentioned in the original caption, such as "a young girl drinking from a pink cup" in the third example.

\subsection{Features Enhancement}
\label{sec:feature_enhancement}
As transformers show great effectiveness in multiple cross-modal tasks, we enhance both the textual and visual encoders of a state-of-the-art interpretable embedding model \cite{wu2023ITV} with pre-trained transformers. Specifically, the textual encoder in \cite{wu2023ITV} consists of bag-of-words (BoW), W2V, and GRU, i.e., $F(x)=BN(FC([$BoW, W2V, biGRU$]))$, where BN and FC are a batch normalization layer and a fully-connected layer. The W2V is pre-trained on English tags of 30 million Flickr images \cite{flicker30k}. We replace the textual encoder with three recent advanced pre-trained text-visual transformers, i.e., CLIP \cite{CLIP}, BLIP-2 \cite{Li2023BLIP2BL}, and imagebind \cite{girdhar2023imagebind}, i.e., $\hat{F}(x)=BN(FC([$CLIP, BLIP-2, imagebind$]))$. Their weights are frozen during training. Similarly, we add these three pre-trained transformers to the visual encoder in \cite{wu2023ITV} and also freeze their weights in the training. In other words, the original encoder $H(x)=BN(FC([$CNNs,biGRU, biGRU-CNN, swinTrans, SlowFast$]))$ is changed to $\hat{H}(x)=BN(FC([$CNNs, biGRU, biGRU-CNN, swinTrans, SlowFast, CLIP, BLIP-2, imagebind$]))$. The weights of CNNs, swinTrans, and SlowFast are frozen as the same as in \cite{wu2023ITV}.

\section{Experiments}
This section first evaluates the newly introduced components on TRECVid AVS datasets \cite{Trecvid2016,V3C1} through ablation studies. We also compare the proposed methods with the state-of-the-art on the MSRVTT \cite{msr-vtt} and TRECVid AVS datasets.

\subsection{Experiment Setting}
\subsubsection{Datasets}
In the pre-training stage, we use the WebVid-GenCap7M for training, and WebVid2M \cite{Bain21_frozenInTime} val set for validation. Following the setting of the existing AVS approaches \cite{w2vvpp,dual_task,wu2023ITV}, we fine-tune the interpretable embedding model on the combination of TGIF \cite{tgif}, MSRVTT \cite{msr-vtt}, and VATEX \cite{VATEX} datasets, validate it on a TRECVid VTT dataset (i.e., tv2016train) \cite{trecvid2006} and test it on TRECVid AVS datasets (i.e., IACC.3 \cite{Trecvid2016}, V3C1 \cite{V3C1}, and V3C2 \cite{V3C1}). IACC.3 dataset has 334k video clips associated with 90 AVS queries across three years. V3C1 and V3C2 have 1 million and 1.4 million video clips, respectively, where the former is linked with 70 AVS queries used in 2019-2021, and the latter consists of 50 AVS queries used in 2022-2023.  

\begin{table*}[t]
\caption{AVS Performance comparison between word-only and multi-word concept banks, and with and without WebVid-genCap7M in pre-training (PreT)} 
\label{Tab:concept_bank}

    \centering
\begin{tabular}{c|ccc|ccc|ccc}
\toprule
        & \multicolumn{3}{c|}{Concept-based search} & \multicolumn{3}{c|}{Embedding-based search} & \multicolumn{3}{c}{Fusion search} \\ \hline
   concept bank type     & word  &word   &  multi-word     & word    &word    &  multi-word      & word   &word     &  multi-word      \\\hline
 with/without pre-training       & w/o PreT  &w/ PreT   &  w/o PreT     & w/o PreT  &w/ PreT   &  w/o PreT       & w/o PreT  &w/ PreT   &  w/o PreT       \\\hline
tv16     
& 0.184      &0.171   & \textbf{0.197}        
& 0.187   &\textbf{0.208}      & 0.179          
& 0.211   &\textbf{0.212}     & 0.209          
\\
tv17     
& 0.230   &0.240 & \textbf{0.288}           
& 0.279   &\textbf{0.290} & 0.283            
& 0.292   &0.302 & \textbf{0.325}         
\\
tv18    
& 0.135          &0.127     & \textbf{0.162}          
& \textbf{0.140} &0.139     & 0.134            
& \textbf{0.170} &0.158     & 0.169        
\\
tv19    
&0.166           &0.167            & \textbf{0.197}         
& 0.201          &\textbf{0.205}   & 0.203         
& \textbf{0.227} &0.216            & 0.222           
\\
tv20    
& \textbf{0.292}  &0.233         & 0.285           
& 0.307           &0.312         & \textbf{0.319}           
& 0.345           &0.321         & \textbf{0.346}          
\\
tv21     
& 0.246           &0.262 & \textbf{0.267}             
& 0.294  &\textbf{0.295}          & 0.284          
&\textbf{0.318}            &\textbf{0.318} & 0.308    
\\
tv22     
& 0.115           &0.090   & \textbf{0.116}             
& \textbf{0.135}  &0.123   & 0.131           
& \textbf{0.150}  &0.119   & 0.149 
\\
tv23     
& 0.124          &0.089  & \textbf{0.186}             
& 0.151          &\textbf{0.160}  & 0.153           
& 0.167          &0.147  & \textbf{0.195} 
\\ \hline
mean  
& 0.186          &0.172           & \textbf{0.212}            
& 0.212          &\textbf{0.216}  & 0.211            
& 0.235          &0.224           & \textbf{0.240}     \\

 \bottomrule  
\end{tabular}
\end{table*}

\begin{table*}[]
\caption{Performance comparison on feature enhancement. $F(x)$ and $H(x)$ are the baseline textual end visual encoders, while  $\hat{F}(x)$ and $\hat{H}(x)$ are  with enhanced textual and visual features.}
\label{tab:feature_enhancement}
\centering
\resizebox{0.98\linewidth}{!}{
\begin{tabular}{c|c|ccc|ccc|cc|c}
    \toprule
&                     & \multicolumn{3}{c|}{IACC.3}               & \multicolumn{3}{c|}{V3C1}   & \multicolumn{2}{c|}{V3C2}   \\
\hline
textual encoder &visual encoder                         & tv16 (30)         & tv17 (30)           & tv18  (30)          & tv19 (30)   &tv20 (20) &tv21 (20) &tv22 (30) &tv23 (20)    &mean   \\ 
    \midrule

$F(x)$ & $H(x)$  & 0.211  & 0.292   &\textbf{0.170}   &  0.227 & 0.345   & 0.318   &  0.150& 0.167 &0.235 \\ 
\hline
$\hat{F}(x)$  & $H(x)$   &  0.216 & 0.288   &  0.148 &  0.204 & 0.325   & 0.307   & 0.150 &  0.173 &0.226 \\ 
$\hat{F}(x)$   &$\hat{H}(x)$   & 0.249   & 0.279   & 0.168  & 0.243  &  0.360  &  0.364  & \textbf{0.215}  &\textbf{ 0.250} &0.266  \\
$F(x)$ &$\hat{H}(x)$   & \textbf{0.254}  &  \textbf{0.318}  & 0.162  & \textbf{0.254 } & \textbf{0.364 }  & \textbf{ 0.368}  & 0.179 & 0.241 &0\textbf{.268}\\
\bottomrule
\end{tabular}
}
\end{table*}

\subsubsection{Evaluation Metric}
We follow the AVS standard \cite{trecvid2006} to report xinfAP with a search length of 1000 on TRECVid AVS datasets. For the text-to-video retrieval on the MSRVTT dataset, we report R@{1,5,10}, MedR, and mAP.

\subsubsection{Implementation Details}
We implement our proposed methods based on the publicly available code provided by the interpretable embedding model (ITV) \cite{wu2023ITV} and use the same parameter setting as ITV. We follow ITV to use cosine similarity to measure the alignment between the query and videos and set equal weight for the concept-based search and embedding-based search in the fusion. For the concept bank extraction, we use Stanford coreNLP parser\footnote{https://nlp.stanford.edu/software/srparser.html} to obtain the parse tree of a sentence. 
For the image caption generation, we use the base and large BLIP models \cite{li2022blip} trained on the MSCOCO dataset \cite{MSCOCO}.

\subsection{Ablation Studies}
The section studies the impact of the proposed three components, and the original interpretable embedding model \cite{wu2023ITV} is used as a baseline.

\subsubsection{Word-only versus Multi-word Concept Banks}
\label{sec:ab:concept_bank}



\begin{figure}[]
    \centering
    \includegraphics[width=\linewidth]{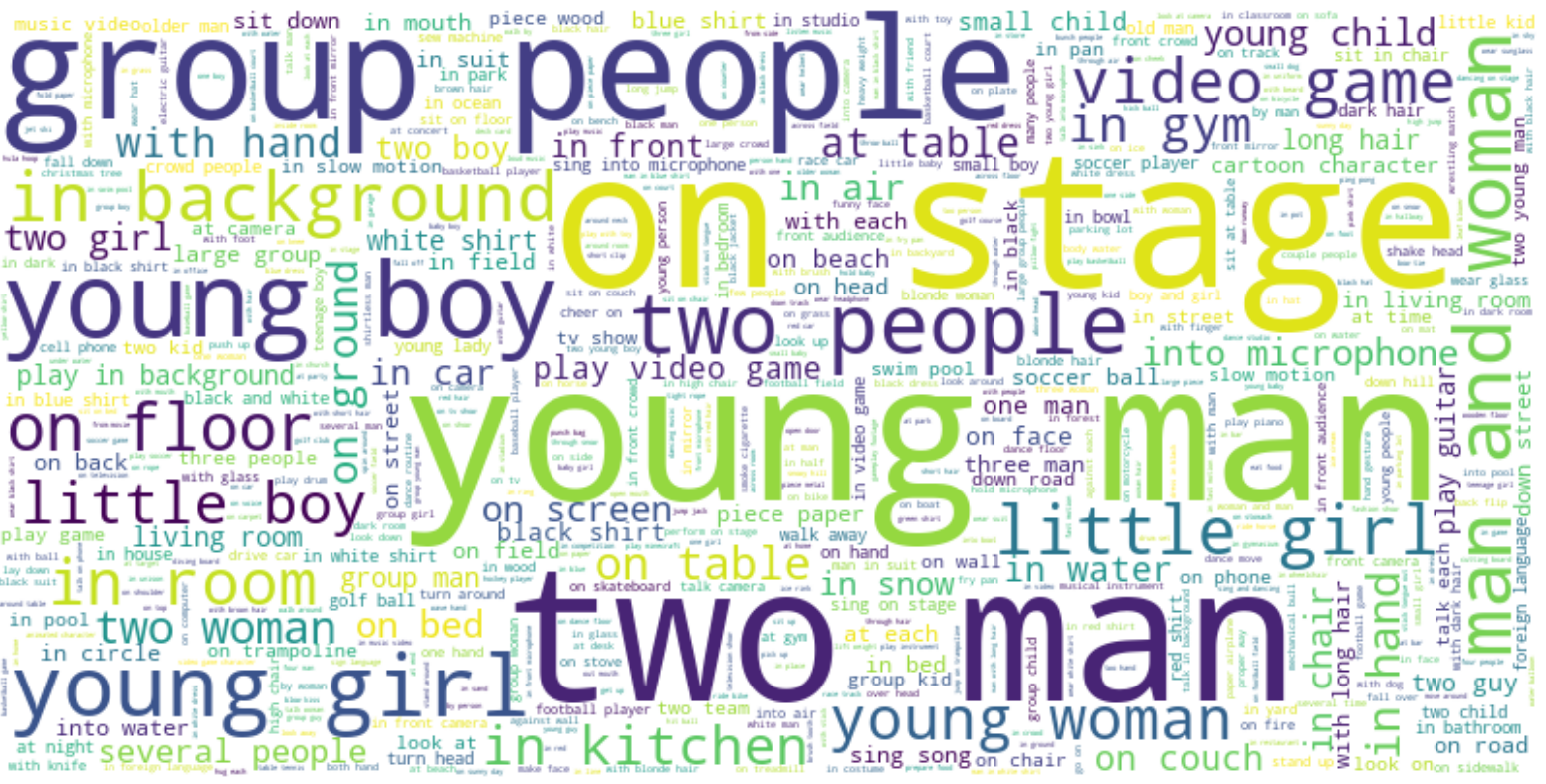}
    \caption{Cloud figure of phrases in the multi-word concept bank.}
    \label{fig:concept_word_cloud}
\end{figure}

Figure \ref{fig:concept_word_cloud} visualizes the phrases and their frequencies in the multi-word concept bank built on a caption corpus. The corpus contains all the video captions of TGIF \cite{tgif}, MSRVTT \cite{msr-vtt}, and VATEX \cite{VATEX} datasets. The concept bank has 14,528 concepts, 9,465 of which are phrases. 62\% of phrases appear between 20 to 50 times, and 18\% appear more than 100 times in the training corpus. As shown in Figure \ref{fig:concept_word_cloud}, the concept bank manages to contain five main types of phrases, including noun phrases such as \textit{man and woman}, verb phrases such as \textit{sit down}, adjective phrases such as  \textit{young man}, prepositional phrases such as \textit{on floor} and quantifier phrases such as \textit{two man}.

We compare the AVS performances with two different concept banks in interpreting embeddings without pre-training. Table \ref{Tab:concept_bank} contrasts the retrieval results on eight query sets across three search modes: concept-based search, embedding-based search, and the fusion of them. Our proposed multi-word concept bank significantly outperforms the word-only concept bank consistently on the concept-based search across most query sets and boosts the concept-based search to be competitive with the embedding-based search interpreted by a word-only concept bank. About 65\% of queries get improved on the concept-based search, and about 57\% of them are bad-performing queries (i.e., xinfAP < 0.1) that suffer from out-of-vocabulary problems. The performance improvement is mainly attributed to a better capability of modeling the relationships between query words. For example, for query-535 \textit{Find shots of a person standing in front of a brick building or wall}, it is almost impossible to use a list of word-only concepts to interpret the position of the person and building/wall. However, with the addition of the prepositional phrase \textit{in front brick wall}, the position relation can be interpreted properly, and the retrieval performance increases by six times. As embedding-based search is good at modeling the relationships of query words, the addition of the phrase concepts does not bring a significant improvement to embedding-based search, and the average retrieval performances of the two concept banks are almost the same. Overall, with the improvement in concept-based search, the multi-word concept bank has a slightly better average xinfAP than the word-only on fusion search.      

\begin{table}[]
\caption{Comparison on the MSRVTT dataset. 
All performances are cited directly from the published results. Except CLIP4CLIP is rerun and tested on the same split of the MSRVTT dataset. The unavailable results are  marked with /.
} 
\label{exp:msrvtt}
\centering
\begin{tabular}{l|ccccc}
\toprule
              & \multicolumn{5}{c}{Text-to-Video Retrieval} \\ \hline
              & R@1    & R@5    & R@10   & Med r   & mAP    \\
              \hline
W2VV++ \cite{w2vvpp}       & 11.1   & 29.6   & 40.5   & 18      & 20.6   \\
Dual encoding \cite{dualconding} & 11.1   & 29.4   & 40.3   & 19      & 20.5   \\
Hybrid space \cite{tpami21-dual-encoding} & 11.6   & 30.3   & 41.3   & 17      & 21.2   \\
Dual-task    &12.2  &  31.7   &  42.9    & 16      &   22.3  \\
SEA \cite{tmm2021-sea}          & 13.1    & 33.4     & 45.0     & 14     & 23.3     \\
RIVRL \cite{RIVRL}         & 13.0    &  33.4    &  44.8    &   14     &  23.2     \\
LAFF \cite{LAFF}          & 16.0    &  39.5    & 51.4     &  10     &   27.6    \\
ITV   \cite{wu2023ITV}      & 13.2   & 33.7   & 45.0   & 14      & 23.4 \\

CLIP4CLIP \cite{Luo2021CLIP4Clip}  &28.6   & 53.2 & 63.8&   5   &  / \\
 \hline 
improved ITV        & \textbf{29.3}   &  \textbf{54.4} &   \textbf{65.0} &    \textbf{4}   &  \textbf{41.2} \\

\bottomrule
\end{tabular}
\end{table}

\subsubsection{Impact of Pre-training on Video-GenText Pairs}
\label{sec:ab:dataset}

We evaluate the impact of the proposed large-scale video-text dataset in pre-training interpretable embeddings. Table \ref{Tab:concept_bank} compares the AVS performance of interpretable embedding with and without pre-training on the WebVid-genCap7M dataset with the word-only concept bank. The pre-trained model has better performances on six out of eight query sets and obtains the highest overall performance on embedding-based search. Specifically, taking advantage of the pre-training, over 50\% of queries have higher xinfAP on embedding-based search. However, on the concept-based search, the pre-training degrades the xinfAPs on five out of eight query sets, and there are considerable drops on the tv22 and tv23 query sets. As observed, these are mainly attributed to the ground truth problem. For example, about 60\% of the video clips retrieved by the pre-trained model on concept-based search on the tv22 query set are newly found and not judged in the evaluation in that year. When computing the xinfAP metric, the unjudged video shots are regarded as wrong results. We compute a xinfAP with a shorter search length (i.e., 10) where there is a smaller percentage of unjudged video clips, the pre-trained model exceeds the model without pre-training with xinfAP@10=0.281 to xinfAP@10=0.278 on the tv22.

\subsubsection{Impact of Advanced Textual and Visual Features}
\label{sec:ab:enhanced_Features}

In this section, we investigate whether the integration of recent advanced textual/visual features to the interpretable embedding model (ITV) \cite{wu2023ITV} improves the retrieval performance. The experiment is conducted on TRECVid AVS datasets with eight query sets. The original setting of the ITV is used as a baseline. We change the textual encoder from $F(x)$ to $\hat{F}(x)$ and the visual encoder from $H(x)$ to $\hat{H}(x)$ in Section \ref{sec:feature_enhancement} to include recent advanced features, i.e., CLIP \cite{CLIP}, BLIP-2 \cite{Li2023BLIP2BL}, and imagebind \cite{girdhar2023imagebind},

Table \ref{tab:feature_enhancement} compares the AVS performances with different combinations of encoders. On the text side, the transformer-based pre-trained models are not comparable to the traditional encoders on most AVS query sets and degrade the mean performance of the baseline by 4\%. In contrast, except for the tv17 query set, the addition of the advanced visual features does boost the retrieval performances consistently by a large margin. It also beats the enhanced features on both text and video sides on most query sets.

\subsection{Text-to-video Retrieval on MSRVTT Dataset}

We compare the proposed model with other existing methods on the MSRVTT dataset \cite{msr-vtt}. The proposed model is with the multi-word concept bank, pre-trained on the WebVid-genCap7M dataset, and with enhanced features on both text and video sides. Table 5 reports the retrieval results on the official split, which has 6513, 497, and 2990 videos for training, validation, and testing, respectively. With all the three proposed components, the improved ITV manages to exceed other approaches significantly. Especially, it outperforms the original ITV \cite{wu:sql} more than two times on R@1 and has higher mAP on 64.2\% test queries. Furthermore, compared to CLIP4CLIP \cite{Luo2021CLIP4Clip}, which uses the pre-trained weights of CLIP and fine-tunes on MSRVTT, the improved ITV also shows better performances on all evaluated metrics.

\subsection{AVS Comparison to the State-of-the-art}

\begin{table*}[]
\caption{Performance comparison across eight years of TRECVid AVS datasets. The number inside parentheses indicates the number of queries evaluated that year. The reproduced results are marked with *} 
\label{tab:tenSearch_avs_comparison}
\centering
\begin{tabular}{l|ccc|ccc|cc}
    \toprule
Datasets                     & \multicolumn{3}{c|}{IACC.3}               & \multicolumn{3}{c|}{V3C1}   & \multicolumn{2}{c}{V3C2}    \\
\hline
Query sets                         & tv16 (30)         & tv17 (30)           & tv18  (30)          & tv19 (30)   &tv20 (20) &tv21 (20) &tv22 (30) &tv23 (20)      \\
                          \hline
\multicolumn{5}{l}{TRECVid top result:}                                                  \\
top-1     & 0.054  & 0.206  & 0.121 & 0.163   &\textbf{0.359}   &0.355   &\textbf{0.282} &0.292\\
\hline
\multicolumn{5}{l}{Pre-trained models:}  \\ 
CLIP  \cite{CLIP}       & 0.182   &  0.217   & 0.089  &0.117   & 0.128  &  0.178  &0.124 & 0.109\ \\
BLIP-2 \cite{Li2023BLIP2BL}     &0.213 &0.226 & 0.168&0.199 &0.222 &0.273 &0.164 & 0.203 \\
CLIP4CLIP \cite{Luo2021CLIP4Clip} & 0.182 &0.217 &0.089 &0.133 & 0.149 &0.188 & 0.121 & 0.109 \\
\hline
\multicolumn{5}{l}{AVS approaches:}  \\  
ConBank  & /  & 0.159 \cite{Waseda_Meisei2017}  & 0.060 \cite{Waseda2018}        & /    & /     & /     & / &/  \\
ConBank (manual)   & 0.177 \cite{Waseda2016}   & 0.216 \cite{Waseda_Meisei2017}  & 0.106 \cite{Waseda2018}        & 0.114  \cite{WasedaMeiseiSoftbank2019}  & 0.183 \cite{wasedaAVS2020}    & /  & /  &/   \\
W2VV++ ~\cite{w2vvpp}       & 0.150         & 0.207           & 0.099         & 0.146     & 0.199      & /  & /  &/     \\
Dual coding ~\cite{dualconding}              & 0.160          & 0.232           & 0.120         & 0.163  & 0.208    & /  & /  &/     \\ 
Dual-task \cite{dual_task}            & 0.184          & 0.252           & 0.120       &0.189    & 0.229  & 0.193  & /  &/ \\ 
HGR ~\cite{hgr}             &/         & /           & /       & 0.142   &0.301     & /  & /  &/    \\ 
SEA ~\cite{tmm2021-sea}             & 0.164         & 0.228           & 0.125       & 0.167     & 0.186  & /  & /  &/   \\

 

Hybrid space ~\cite{tpami21-dual-encoding}              & 0.157          & 0.236           & 0.128         & 0.170    & 0.191  & 0.162 & /  &/    \\

LAFF* \cite{LAFF}             &    0.188   &   0.261        &  0.152   &  0.215     &  0.299   &  0.300  & 0.178 &0.172  \\ 
RIVRL*~\cite{RIVRL}  & 0.159          & 0.231           & 0.131        & 0.197    & 0.278  & 0.254 & 0.179  & 0.177  \\ \hline
ITV$_{concept}$  \cite{wu2023ITV}              & 0.184     &  0.230    &  0.135  &   0.166          &0.292  & 0.246 & 0.115 & 0.124 \\
ITV$_{embedding}$  \cite{wu2023ITV}              & 0.187     &  0.279    &  0.140  &   0.201          &0.307  & 0.294 & 0.135 & 0.151 \\
ITV$_{fusion}$  \cite{wu2023ITV}              & 0.211     &  0.292    &  \textbf{0.170}  &   0.227          &0.345  & 0.318 & 0.150 & 0.167 \\
\hline
\multicolumn{5}{l}{The proposed models:}  \\
improved ITV$_{concept}$               & 0.252     &  0.310    &  0.127  &   0.161          &0.245  & 0.295 & 0.164 & 0.280 \\
improved ITV$_{embedding}$               & 0.233     &  0.296    &  0.167  &   0.237          &0.334  & 0.309 & 0.198 & 0.241 \\
improved ITV$_{fusion}$               & \textbf{0.280}     &  \textbf{0.349}    &  0.165  &   \textbf{0.242}          &0.352  & \textbf{0.365} & 0.235 & \textbf{0.295} \\

\bottomrule
\end{tabular}
\end{table*}

Table \ref{tab:tenSearch_avs_comparison} compares our proposed method with widely used pre-trained models, existing AVS approaches, and the top-1 performances reported by TRECVid in each year of evaluation. 
For the recent approaches, LAFF \cite{LAFF} and RIVRL \cite{RIVRL}, we reproduce their results using the same setting as ours for fair comparisons (e.g., the same training sets and visual features) based on their published codes and report the results across eight years of query sets (tv16 to tv23).

\begin{figure*}[]
    \centering
    \includegraphics[width=\linewidth]{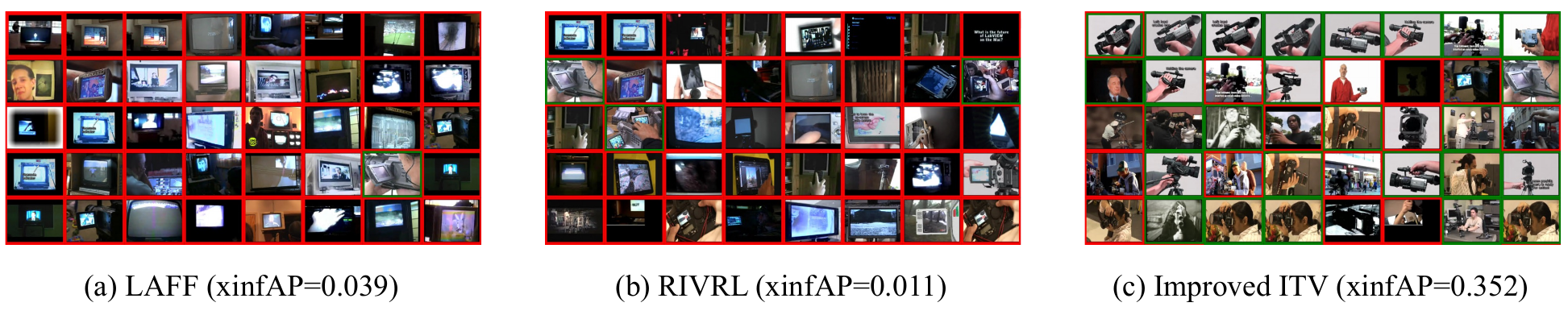}
    \caption{Comparison with the state-of-the-art approaches LAFF and RIVRL on query-554 \textit{Find shots of a person holding or operating a tv or movie camera}.}
    \label{fig:comparison_laff_RIVRL}
\end{figure*}

As shown in Table \ref{tab:tenSearch_avs_comparison}, our proposed improved ITV significantly outperforms the pre-trained models on large image-text datasets and all the existing AVS approaches by a large margin. Specifically, compared to the most recent AVS approaches, LAFF and RIVRL, our model has better performances on 74.3\% and 80\% of queries, respectively. It also manages to improve 31.5\% of queries that LAFF and RIVRL both have xinfAP lower than 0.1 to over 0.1. Figure \ref{fig:comparison_laff_RIVRL} shows an example contrasting LAFF and RIVRL on the query-554. As a TV or movie camera is not frequently seen in video caption datasets, LAFF and RIVRL find video shots of TV instead. With the large-scale pre-trained WebVid-genCap7M, the improved ITV outperforms them significantly, even though some correct results are not annotated(judged) in the ground truth. Furthermore, the improved ITV also outperforms or is competitive with the top-1 solution on most query sets even though the top-performance teams in TRECVid AVS evaluation usually fuse multiple rank lists from various models, such as the top-1 solution on the tv22 query set fusing the results of more than 100 rank lists from five models. 

The performance improvements over the original ITV are consistent across concept-based, embedding-based, and fusion searches. There are 139, 136, and 134 out of 210 queries having higher performances on the three search modes, respectively. The number of queries whose xinfAP are less than 0.1 also decreases by 13\%, 31\%, and 27\% on three search modes, respectively. This is mainly due to the effectiveness of the three proposed components in addressing the out-of-vocabulary problem suffered by these queries. The three proposed components are complementary as their combination outperforms the performances of all three individual components as shown in ablation studies.

\section{Conclusion}

In this paper, we study three components to address the small-size dataset and out-of-vocabulary problems on the AVS task. The multi-word concept bank boosts the performance of out-of-vocabulary queries by enhancing the capability of concept-based search on modeling relationships between query words. The newly constructed video-GenText dataset manages to improve the embedding-based search by having more training instances on unseen queries. The recent-advanced visual features manage to increase the retrieval performance, while the advanced textual features are not competitive with the traditional features on the AVS task. The three introduced elements are shown to be complementary, and their combination significantly increases the retrieval performances of an interpretable embedding model and outperforms the state-of-the-art AVS approaches on both small (i.e., MSRVTT) and large datasets (i.e., TRECVid AVS datasets).

\bibliographystyle{ACM-Reference-Format}

\bibliography{AVS}
\end{document}